\documentclass[preprint,12pt,authoryear]{elsarticle}

\usepackage{amssymb}
\usepackage{amsthm}

\usepackage{hyperref}
\usepackage{booktabs}
\usepackage{multirow}
\usepackage{tikz}
\usepackage[normalem]{ulem}

\journal{arXiv}

\begin{document}

\begin{frontmatter}



\title{Towards Language-driven Scientific AI}


\author[label1]{José Manuel Gómez-Pérez}

\affiliation[label1]{organization={Language Technology Research Lab, Expert.ai},%Department and Organization
            addressline={3 Poeta Joan Maragall}, 
            city={Madrid},
            postcode={28020}, 
            %state={},
            country={Spain}}

\begin{abstract}
Inspired by recent and revolutionary developments in AI, particularly in language understanding and generation, we set about designing AI systems that are able to address complex scientific tasks that challenge human capabilities to make new discoveries. Central to our approach is the notion of natural language as core representation, reasoning, and exchange format between scientific AI and human scientists. In this paper, we identify and discuss some of the main research challenges to accomplish such vision. 
\end{abstract}


%\item This paper presents a methodological framework based on artificial intelligence and natural language processing and understanding to support text analytics and machine understanding tasks in space, providing in addition a series of recommendations and lessons learnt. 
%\item The paper also presents several case studies illustrating the application of such framework to solve real-life information extraction and machine understanding needs across a rich representation of different functional areas at ESA. In doing so, we address complex information extraction and language understanding challenges that had not been addressed in space until now.

\begin{keyword}
Science \sep Artificial Intelligence \sep Language Understanding


\end{keyword}

\end{frontmatter}


\section{Introduction}
\label{sec:intro}
During her presidential address at the AAAI Conference,~\cite{Gil_2022} pondered whether artificial intelligence (AI) will write scientific papers in the future. She believed that we can be hopeful that the answer is yes and that it may happen sooner than we might expect. As scientific questions become significantly more complex, our capabilities to do scientific breakthroughs need to be augmented. Compare for instance the challenges of formulating Kepler's laws of planetary motion or the discovery of a cure for Polio with demonstrating the existence of binary stellar-mass black hole systems~\citep{PhysRevLett.116.061102} or the treatment of glioblastoma, a type of brain cancer. While the former were achieved by a single scientist, the latter require large and interdisciplinary teams involving the collaboration of hundreds of scientists from different fields to work together during years to produce results. 

In this paper, we present a personal perspective inspired by recent breakthroughs in AI and particularly language technologies to enable a next generation of AI systems that may become an effective part of the scientific ecosystem, collaborate, contribute, and eventually produce significant findings~\citep{Kitano_2016}. In recent years, the incorporation of intelligent techniques for data mining and machine learning has provided scientists with powerful data-driven analytics and discovery capabilities. However, such techniques have been focused on solving well-defined narrow tasks. Confining intelligent machines to such tasks can severely limit our ability to truly harness the potential of AI to enable us to tackle larger scientific problems. 

It is time to take a quantum leap. Future scientific endeavors will require partnerships of scientists and AI, where machines may independently pursue substantial aspects of the research and contribute their own discoveries. Such thoughtful AI systems~\citep{gil-ds17} should be capable of formulating their own research goals, proposing and evaluating hypotheses, designing theories, debating alternative options, and generating new knowledge. They should be able to explain their reasoning, compare their rationales to others, and situate their findings in the existing literature. AI systems should be able to communicate with scientists with different levels of expertise in a topic. To form a true partnership, they should be able to take guidance from scientists as well as to provide guidance to them. Today, this vision is still impossible to the point that new research is required to make it happen. 

The following sections delve into the challenges this vision entails and how it could be accomplished from a language-driven research perspective.

\section{Scientific AI will be language-driven}
As part of the scientific task forces of the future, AI systems will need to exchange feedback with human scientists and learn from their interaction. Rather than fixed, structured formalisms to represent scientific knowledge, which can be brittle and constrained to our ability to represent things explicitly, we propose a natural language-driven approach where language is the main formalism to represent and exchange scientific information between the different agents in the scientific ecosystem, be they humans or machines. 

Generative language models like GPT-3~\citep{brown2020language} or T5~\citep{raffelT52020} produce realistic human text based on a statistical bias acquired through self-supervised training over an extremely large document corpus, learning to guess the word that is most likely to come next given a prompt, with applications in many language tasks like information extraction, reading comprehension and question answering, conversation, summarization or machine translation. Such models promote a change of paradigm in NLP, from “pre-train, fine-tune, predict” to “pre-train, prompt, predict”, where a prompt is a piece of text inserted in the input examples so that the task that needs to be solved can be formulated as a language modeling problem. Subsequently, prompt-based prediction~\citep{gao-etal-2021-making, schick-schutze-2021-exploiting} seeks to specify such prompts as effectively as possible. 

We posit that the task of formulating research goals, hypotheses, and claims by machines in natural language, as well as the evaluation of those produced by other scientists, can be recast into a series of instructions and prompts in natural language that inform the model. However, there is no research that has explored this path yet. Generative language models and prompt-based prediction are promising but still in their infancy, scientific tasks like the formulation of hypotheses, goals and claims require a level of knowledge, abstract thinking and reasoning only humans have been capable of yet, and there are no datasets that enable the evaluation and testing of systems that aim to solve such tasks at human level. 

\section{...but also multi-modal}
Although we propose language as the main representation and exchange formalism for scientific AI systems, scientific knowledge is heterogeneous and can present itself in many forms. As originally put by~\cite{Reddy_1988}, \textit{"Reading a chapter in a college freshman text and answering the questions at the end of the chapter is a hard problem that requires advances in vision, language, problem-solving, and learning theory.”}. As of today, this is still one of the grand challenges to be tackled in AI. 

Like many other manifestations of human thought, scientific discourse usually adopts the form of a narrative, a scientific publication or technical report, where related information is presented in mutually supportive ways over different modalities, including text, diagrams, figures, mathematical equations or tables, which need to be accounted for, represented, and understood across the different modalities. Visually grounded language and visual reasoning is frequent in Science. However, dealing with scientific visual information entails additional complexity compared to natural images. 

For example, scientific diagrams are more abstract and symbolic than natural images, hindering the application of conventional language and vision understanding methods. Some approaches like~\citep{kembhavi2016} propose to parse diagram components and connectors as a graph that can be subsequently interpreted. However, this approach does not seem to generalize beyond a few types of predefined diagrams (water cycle, food chain, etc.). Others~\citep{gomezperez2019} leverage the free supervision provided by the correspondence between scientific figures and the text in their captions to generate a unified language-vision representation space that enables language-vision understanding. Experimental results showed the emergence of visual representations capturing certain features, such as line plots, whisker plots or immunoblots, and their combination in more complex figures. Follow up work by~\cite{gomez-perez-ortega-2020-isaaq} tap on language models and cross-modal attention to identify regions of interest corresponding to diagram components and their relationships to improve the selection of relevant visual information in order to answer different types of questions. 

Unlike natural image datasets for visual question answering or image captioning like COCO~\citep{cocodataset} and Visual Genome~\citep{krishna2017}, there are barely any datasets that are rich with annotated scientific diagrams or such dataset are too small to train large models (AI2D\footnote{\url{https://allenai.org/data/diagrams}}). So far, this has been a strong limitation for any significant progress in this area that our work aims to address.

\section{Generating problem-solving strategies}
Tackling scientific problems of such complexity that may challenge human scientists will require AI systems to learn strategies (methods) that allow decomposing the scientific task at hand into simpler, more attainable steps, following a divide-and-conquer approach. The notion of problem-solving methods~\citep{Mcdermott1988} was originally proposed in the context of expert system research and then applied to answer scientific questions in disciplines like chemistry, physics, and biology~\citep{GOMEZPEREZ2010641}. However, while the resulting systems were able to provide effective strategies to answer certain types of scientific questions, such “recipes'' tend to be rigid, brittle, and hard to generalize. This problem can be seen as an instance of the knowledge acquisition bottleneck~\citep{feigenbaum1984}, where the resulting model suffers from the cognitive limitations that humans may experience to identify, formulate, and explicitly represent the potentially vast number of possible cases to be covered. 

Therefore, rather than exclusively modeled by experts, scientific problem-solving methods should also be learnt from the data. Language models have shown good results in tasks like multi-hop question answering~\citep{Mavi2022ASO}, where answering a question requires several steps. Recent work~\citep{Wei2022ChainOT} explores the ability of language models to generate a coherent chain of thought as a series of short sentences that mimic the reasoning process a person might follow when responding to a question. Indeed, inducing a chain of thought via prompting has shown to enable sufficiently large language models to better perform reasoning tasks. Even more recently, other approaches~\citep{Khot2022DecomposedPA} advocate for solving complex tasks by directly decomposing such tasks into simpler sub-tasks via prompting, optimizing each prompt for its speciﬁc task, further decomposing and, if necessary, replacing the prompt with more effective ones, trained models, or symbolic functions. 

\section{Factual, argumentative, and ethical scientific AI}
It is common for generative language models like GPT-3 to produce text that is realistic but also hallucinatory or nonsensical.\footnote{GPT-3, Bloviator: OpenAI’s language generator has no idea what it’s talking about \\ \url{https://www.technologyreview.com/2020/08/22/1007539/gpt3-openai-language-generator-artificial-intelligence-ai-opinion}} In science we need models that do not just look thoughtful and able to reason with scientific information, but models that as a matter of fact are scientific. Tasks like entailment have been used with good results~\citep{pasunuru-bansal-2018-multi} in support of generative tasks like abstractive text summarization to ensure coherence between the summary and the original text, increasing factuality. However, science also requires the ability to link the scientific hypotheses, goals, and claims generated by the model with actual evidence that supports such statements. Just like human scientists produce new research based on previous work and cite such work to sustain their research, scientific AI will need to contrast their research goals, hypotheses, and claims with evidence. 

Previous work in this direction, like SciFact~\citep{wadden-etal-2020-fact}, explores the validation of claims against the literature to determine whether they are supported or refuted by previous work. We plan to take a step further, creating an evidence-based framework for credibility review inspired by recent breakthroughs in fact-checking and misinformation detection, like the acred credibility review framework~\citep{denaux2020}, both evidence-based, explainable and differentiable. 

On the other hand, future scientific AIs will also require the ability to be argumentative and justify their own reasoning in natural language, driven by a notion of reward that stems from winning a scientific discussion and positively reinforces such behavior. Furthermore, scientific AI will need to act in ways that are compliant with human best practices and principles. This requires the AI to be self-aware, including a quantifiable notion of proficiency in each particular topic, as well as explainable, faithful and truthful. In this regard, approaches like~\citep{Dalvi2022TowardsTR} generate chains of reasoning that show how the answers to questions are implied by the model’s own internal beliefs, allowing users to interact with the explanations to identify erroneous beliefs and provide corrections.

\section{Modularity, adaptation and autonomous learning}
We need large models that contain representations of vast amounts of knowledge in core and progressively capilar scientific disciplines, with a grounding in world knowledge and commonsense understanding, as well as the ability to continuously acquire and update the model's beliefs as the state of the art evolves. Until now, scientific language models like SciBert~\citep{beltagy-etal-2019-scibert}, BioBert~\citep{lee2019} or SpaceRoBERTa~\citep{9548078} have tried to address the challenge of domain-specifty through additional pre-training on large amounts of scientific documents that leverage large-scale open access scientific resources like OpenAire\footnote{\url{https://www.openaire.eu}}, arXiv\footnote{\url{https://arxiv.org}}, Web of Science\footnote{\url{https://clarivate.com/webofsciencegroup/solutions/web-of-science}} or Semantic Scholar.\footnote{\url{https://www.semanticscholar.org}} However, only such knowledge which is statistically significant in the training data is effectively captured, limiting the usefulness of pre-trained language models as knowledge bases or reasoning engines. Domain adaptation based on additional pre-training may also entail generality loss, impacting on downstream tasks~\citep{garcia-silva-2022a}.

Large language models have recently been shown to generate more factual responses by employing modularity in combination with retrieval~\citep{Adolphs2021ReasonFT, zhou-etal-2022-think}, starting to include internet search as a way to fill in gaps in their internal representations. Science is a moving target, with an exponential annual production of scientific publications\footnote{Science and Engineering publication output continues to grow on average at nearly 4\% per year; from 2008 to 2018, output grew from 1.8 million to 2.6 million articles. In 2018, China (with a share of 21\%) and the United States (with a share of 17\%) were the largest producers. As a group, the EU countries (with a share of 24\%) produced more articles than China or the United States. \url{https://ncses.nsf.gov/pubs/nsb20206/}} that reflects a constant progress of the state of the art. Thus, it is imperative for scientific AI to reach out for knowledge outside of their current beliefs proactively, as a researcher would, in order to have an up-to-date view of the field of interest. Acquiring a quantifiable understanding of how statistically significant the problem to be addressed can be for the internal representations of a scientific language model is needed in order to estimate the type and amount of external knowledge to be brought on board. Defining methods to effectively query and retrieve such knowledge would come next. Finally, more research is required that goes beyond current methods~\citep{peters-etal-2019-knowledge, wang-etal-2021-k} to validate the acquired scientific knowledge, inject such knowledge into the model, and update the model’s beliefs on demand and possibly in real-time.

\section{Conclusions}
\label{sec:conc}
In the future, scientific AI systems should be able to deal with questions like \textit{“What if liver cancer patients are treated with ifosfamide after treatment with trabectedin? and the other way around?”}, \textit{“What if the COVID-19 virus becomes 9\% more infectious?”}, \textit{“What will be the consequences of noise associated with human activities in the Venice lagoon ecosystem?”}. Inspired by recent and revolutionary developments in AI, particularly in language understanding and generation, we set about designing AI systems that are able to propose their own strategies to address complex scientific tasks, including answering such questions and justifying their answers, as well as generating and evaluating research goals, hypotheses, and claims. Central to our approach is the notion of natural language as core representation, reasoning, and exchange format between AI systems and human scientists. Rather than providing an exhaustive list, in this paper we focus on some of the key research challenges that need to be addressed to accomplish such vision.

%\section*{Acknowledgements}



\bibliographystyle{elsarticle-harv} 
\bibliography{towards_sciAI}


%%%\begin{thebibliography}{00}


%%%\bibitem[ ()]{}

%%%\end{thebibliography}
\end{document}